\documentclass[runningheads,a4paper]{llncs}

\usepackage[utf8]{inputenc}
\usepackage{url}
\usepackage{xcolor}
\usepackage{xspace}
\usepackage{amsmath}
\usepackage[colorlinks=true,allcolors=blue]{hyperref}
\usepackage{amssymb}
\usepackage{graphicx} 
\usepackage{caption,subcaption}
\usepackage{verbatim}

\DeclareMathOperator{\enc}{enc}
\DeclareMathOperator{\emb}{emb}
\DeclareMathOperator{\gcn}{gcn}

\newcommand{\keywords}[1]{%
                \par\addvspace\baselineskip%
                \noindent{\bf Keywords.}\enspace\ignorespaces#1}
\makeatletter                
\newcommand*{\rom}[1]{\expandafter\@slowromancap\romannumeral #1@}
\makeatother
\newcommand{\red}[1]{{\color{red}{#1}}}

\newcommand{\mesh}{MeSH\!\textsuperscript{\textregistered}\xspace}
\newcommand{\omim}{OMIM\!\textsuperscript{\textregistered}\xspace}
\newcommand{\medline}{PubMed\!\textsuperscript{\textregistered}\xspace}
\newcommand{\nodeTvec}{\emph{node2vec}\xspace}

\newcommand{\myparagraph}[1]{\vspace{0.4cm}\noindent\textbf{#1}~}

\title{Disease Normalization with Graph Embeddings}
\author{D. Pujary\inst{1,2}, C. Thorne\inst{2} \and W. Aziz\inst{1} 
}
\institute{
\begin{tabular}{@{}c@{}}
University of Amsterdam \quad \and Elsevier\\
\end{tabular}\\
            \url{drv022@gmail.com} \quad
            \url{c.thorne.1@elsevier.com} \quad 
            \url{w.aziz@uva.nl}
}

\begin{document}

\maketitle

\begin{abstract}
The detection and normalization of diseases in biomedical texts are key biomedical natural language processing tasks. Disease names need not only be identified, but also normalized or linked to clinical taxonomies describing diseases such as \mesh. In this paper we describe deep learning methods that tackle both tasks. We train and test our methods on the known NCBI disease benchmark corpus. We propose to represent disease names by leveraging \mesh's graphical structure together with the lexical information available in the taxonomy using graph embeddings. We also show that combining neural named entity recognition models with our graph-based entity linking methods via multitask learning leads to improved disease recognition in the NCBI corpus.
\end{abstract}

\section{Introduction}

In biomedical search applications, e.g. clinical search applications, it is of key importance to query not simply by keywords but rather by \emph{concept}, viz., resolve names into their synonyms and return all matching documents. This is particularly true regarding diseases \cite{Dogan2012AnIM}. When we issue a query about ``ovarian cancer'', we expect the system to return hits mentioning standard synonyms such as ``ovarian neoplasms''. To achieve this it is necessary not only to perform \emph{named entity recognition} (NER) to detect all entities mentioned in a document, but also to disambiguate them against databases of canonical names and synonyms, a task known as \emph{entity normalization} or \emph{linking} (EL).

Detecting and normalizing disease mentions are challenging tasks due to linguistic variability (e.g. abbreviations, morphological and orthographic variation, word order) \cite{10.1093/bioinformatics/btt474,DBLP:conf/nldb/ThorneK18}. Interest in these tasks led to the creation of disease knowledge bases and annotated corpora. The \emph{Medical Subject Headings} (\mesh) taxonomy \cite{MeSH2000} is a repository of clinically-relevant terms covering (among others) a large range of standardized disease names, where standard names and known synonyms are organised hierarchically. The NCBI disease corpus \cite{Dogan:2014:NDC:2772763.2772800} is a collection of \medline abstracts with disease mentions manually resolved to \mesh or \omim concepts. 
 
Annotated resources allow the training of supervised learning algorithms. The first such system was \cite{10.1093/bioinformatics/btt474}, which used a learning-to-rank approach. More recently, systems based on deep learning have been proposed \cite{CNN_ranking_EL_2017,Cho2017,2019normco}. The state-of-the-art for NER employs bidirectional recurrent neural networks to parameterize a conditional random field \cite{DBLP:journals/corr/LampleBSKD16}. Normalization is typically based on learning a similarity metric between disease mentions and their standard names in \mesh, where mentions and names are represented by a composition of word embeddings \cite{CNN_ranking_EL_2017,Cho2017}. This typically requires first identifying potential candidate diseases, for example by exploiting handcrafted rules \cite{CNN_ranking_EL_2017} or bidirectional recurrent neural networks \cite{DBLP:conf/emnlp/GreenbergBVM18}. While \cite{2019normco} leverages in addition a coherence model that enforces a form of joint disambiguation. Crucially, previous work has consistently ignored the hierarchical structure of \mesh turning the taxonomy into a flat list of concepts.

In this paper we study an alternative approach to disease normalization. In particular, we acknowledge the hierarchical structure of the \mesh and employ graph encoders---both supervised and unsupervised---to learn representations of disease names (i.e. \mesh nodes). To make node representations easier to correlate to textual input, we propose to also exploit lexical information available in scope notes. Finally, given the similarities between NER and EL, we explore multitask learning \cite[MTL;][]{Caruana:1997:ML:262868.262872} via a shared encoder. MTL has been traditionally applied in biomedical NLP to solve NER over several, though related corpora, covering different, but related entities \cite{10.1093/bioinformatics/bty869}, but has seldom been used to learn related tasks over the same corpus and entities. Our findings suggest that using lexicalized node embeddings improves EL performance, while MTL allows to transfer knowledge from (graph-embedding) EL to NER, with performance topping comparable approaches for the NCBI corpus \cite{Maryam2017deep,10.1093/bioinformatics/bty869}.

\section{Datasets and Methods}

\begin{figure}[tb]
    \centering
    \begin{tabular}{|p{\textwidth}|}
    \hline
    Identification of APC2, a homologue of the $\red{[}$\emph{adenomatous polyposis coli tumour}$\red{]}_{\red{\text{D011125}}}$ suppressor.\\
    \hline
    \end{tabular}\\
    \vspace{0.2cm}
    \begin{tabular}{|p{\textwidth}|}
    \hline
    \textbf{MeSH Heading} \,\quad Adenomatous Polyposis Coli\\
    \textbf{Scope Note} \quad\qquad A polyposis syndrome due to an autosomal dominant mutation of the APC\\ 
    \hspace{2.5cm}genes (GENES, APC) on CHROMOSOME 5. (\dots)\\
    \textbf{Tree Number(s)}  \quad C04.557.470.035.215.100 (\dots)\\
    \textbf{Entry Term(s)}  ~~\,\quad Polyposis Syndrome, Familial (\dots)\\
    \hline
    \end{tabular}    
    \caption{(Top) An NCBI corpus excerpt: title of \medline abstract 10021369. In italics, the mention, in red the \mesh identifier. (Bottom) Snapshot of \mesh entry for disease concept D011125. The ``Scope Note'' is a short (10-20 tokens) definition of the concept.}
    \label{fig:NCBI-example}
\end{figure}

\myparagraph{Knowledge Base}
%
\mesh is a bio-medical controlled vocabulary produced by the National Library of Medicine \cite{MeSH2000} covering a large number of health concepts, including $10,923$ disease concepts, and is used for indexing, cataloging and searching bio-medical and health-related documents and information. There are \mesh records of several types divided further into several categories. In this paper we consider only category C \emph{Descriptors}, in other words on \emph{diseases} \mesh disease nodes are arrayed hierarchically (in the form of an acyclic directed graph or tree) from most generic to most specific in up to thirteen hierarchical levels, and have many attributes, such as,  \textit{MeSH Heading}, \textit{Tree Number(s)}, \textit{Unique ID}, \textit{Scope Note} and \textit{Entry Term(s)}. See Figure~\ref{fig:NCBI-example} (bottom)\footnote{Full \mesh example: \url{https://meshb.nlm.nih.gov/record/ui?ui=D011125}}. 

\myparagraph{Supervision}
%
We use the NCBI disease corpus \cite{Dogan:2014:NDC:2772763.2772800} for training and validating our methods. It consists of $792$\footnote{We drop one training abstract because it is repeated: abstract 8528200.} \medline abstracts separated into training, validation and test subsets (see Table \ref{tab:NCBI_table1}). The corpus is annotated with disease mentions mapped to either \mesh or \omim \cite{10.1093/nar/gki033} (whose identifiers we convert to \mesh identifiers if possible using CTD \cite{10.1093/nar/gky868}), Figure \ref{fig:NCBI-example} (top) illustrates an annotated instance in the corpus. 

\begin{table}[tb]
    \caption{NCBI disease corpus statistics.}
    \centering
    \begin{tabular}{|p{0.165\textwidth}|c|c|c|c|c|}
        \hline
        Split  &  Abstracts & Total entities & Unique entities & Unique concept IDs & Tokens\\ \hline
        Training &  $592$ & $5,134$ & $1,691$ & $657$ & $136,088$\\
        Validation & $100$ & $787$ & $363$ & $173$ & $23,969$\\
        Test & $100$ & $960$ & $424$ & $201$ & $24,497$\\ 
        \hline
    \end{tabular}
    \label{tab:NCBI_table1}
\end{table}

\begin{figure}[t]
    \centering
    \includegraphics[scale=0.5]{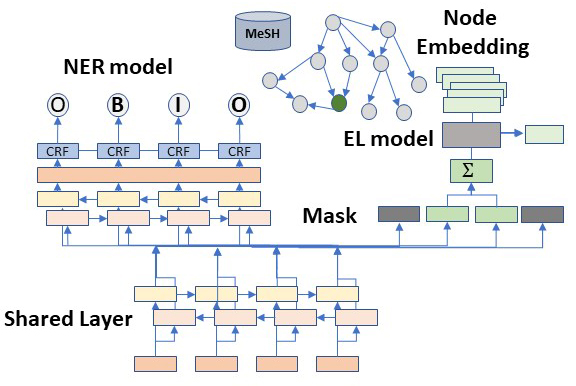}
    \caption{Visualization of the MTL architecture.}
    \label{fig:mtl_arch}
\end{figure}{}

\myparagraph{NER}
%
For NER we use variants of \cite{DBLP:journals/corr/LampleBSKD16} and \cite{DBLP:journals/corr/MaH16} biLSTM-CRF models. First, we encode words in an abstract using either a fixed and pre-trained contextualized encoder, i.e. bioELMo \cite{jin2019probing}, or a concatenation of a fixed and pre-trained biomedical embedding, i.e. SkipGram  \cite{Pyysalo:2013b}, and a trainable character-level encoder (e.g. LSTM \cite{DBLP:journals/corr/LampleBSKD16} or CNN \cite{DBLP:journals/corr/MaH16}). We then re-encode every token with a trainable biLSTM whose outputs are linearly transformed to the potentials of a linear-chain CRF \cite{Lafferty:2001:CRF:645530.655813}. All trainable parameters are optimized to maximize the likelihood of observations (i.e. IOB-annotated spans). At prediction time, Viterbi decoding finds the optimum IOB sequence in linear time.

\myparagraph{EL with \nodeTvec embeddings} 
%
We want to capitalize on the structure of the disease branch of \mesh. Thus, for every disease node, we learn an embedding with \nodeTvec \cite{Perozzi:2014:DOL:2623330.2623732, Grover:2016:NSF:2939672.2939754}. The algorithm starts from randomly initialized node embeddings which are then optimized to be predictive of neighbourhood in the \mesh graph (for this we ignore the directionality inherit to relations in \mesh). We use the \nodeTvec framework \cite{Grover:2016:NSF:2939672.2939754} to generate two types of representations. For \textit{type-}\rom{1}, we use \nodeTvec as is, that is, with randomly initialized node embeddings. For \textit{type-}\rom{2}, we leverage lexical information by initializing a node's embedding with the bioELMo encoding of that node's \emph{scope note} (we use average pooling). The idea behind \textit{type-}\rom{2} embeddings is to combine lexical features in the scope note and  structural relations in \mesh. Our EL model is a logistic regression classifier parameterized by a bilinear product between the embedding $\emb(y)$ of a node $y$ and the contextualized encoding $\enc(x,m)$ of a mention $m$ in an abstract $x$, i.e. $p(y|x,m, \theta) \propto \exp(\enc(x,m)^\top \mathbf W \emb(y))$, where $\theta = \{\mathbf W\}$ are trainable parameters. The node embedding is either a \textit{type-}\rom{1} or \textit{type-}\rom{2} embedding, and the mention embedding is the average of the bioELMo contextualized representation of the tokens in the disease mention. Unlike previous work, we normalize $p(y|x,m, \theta)$ against the set of all known entities in \mesh, that is, without a pre-selection of candidates. The model is optimized to maximize the likelihood of  observations---triples $(m, x, y)$ of mentions in an abstract resolved to \mesh nodes.

\myparagraph{EL with GCN encoders}
%
Graph convolution networks (GCNs) \cite{DBLP:journals/corr/KipfW16} generalize CNNs to general graphs. A GCN is an encoder where node representations are updated in parallel by transforming messages from adjacent nodes in a graph structure. A message is nothing but a parameterized (possibly non-linear) transformation of the representation of the neighbour. Stacking $L$ GCN layers allows information to flow from nodes as far as $L$ hops away in graph structure. In particular we follow \cite{DBLP:journals/corr/MarcheggianiT17}, where the encoding $\mathbf h_v^{(\ell)}$ of a node $v$ in the taxonomy is computed as $\mathbf{h}_v^{(\ell)} = \sigma \left( \sum_{u \in \mathcal{N}(v)}\mathbf{W}^{(\ell)}\mathbf{h}_u^{(\ell-1)} + \mathbf b^{(\ell)} \right)$, where $\ell$ indexes the layer, $\mathcal{N}(v)$ are neighbors of $v$ in \mesh (we again discard directionality), $\mathbf b^{(\ell)}$ and $\mathbf W^{(\ell)}$ are the parameters of the $\ell$th GCN layer, $\sigma$ is a nonlinearity (we use ReLU), and $\mathbf{h}_v^{(0)}=\emb(v)$ is the initial embedding of the disease node $v$. For the $0$th representation of nodes,  we use the bioELMo encoding of a node's scope note. As a trainable encoder, a GCN layer is an integral part of the classifier and its parameters are updated by backpropagation from the downstream objective. Once again, our objective is to maximize the likelihood of observations under a probabilistic classifier, i.e. $p(y|x,m, \theta) \propto \exp(\enc(x, m)^\top \gcn(y; \theta))$, where $\gcn(y; \theta) = \mathbf h_y^{(L)}$ and $\theta = \{(\mathbf W^{(\ell)}, \mathbf b^{(\ell)})_{\ell=1}^L\}$ are the trainable parameters. Note that unlike \nodeTvec embeddings, GCN encoders are trained directly on EL supervision, which is arguably of limited availability.

\myparagraph{Multitask Learning}
%
Traditionally, the problem of disease normalization is tackled by first identifying the disease names (NER) and then normalizing them (EL). We attempt to learn from both types of supervision by having a NER and an EL model share parts of their architectures. This is known as \emph{multitask learning} \cite{Caruana:1997:ML:262868.262872}. In particular, we share the encoder of the NER architecture, see Figure \ref{fig:mtl_arch}, and derive mention features for the EL model from there.

\section{Experiments and Results}

\myparagraph{Hyperparameters}
%
We train our models using ADAM \cite{Adam} and initial learning rate of $10^{-3}$. We split our abstracts into sentences using NLTK\footnote{\url{https://www.nltk.org/}} and process sentences independently. We use $200$-dimensional \emph{word2vec}  embeddings and $1,024$-dimensional bioELMo encodings.  For NER models, we learn $60$-dimensional character embeddings. We train \mesh node embeddings using \nodeTvec for $100$ epochs, whereas GCN-based EL models are trained for $500$ epochs. All models employ $1,024$-dimensional node embeddings. We stack $L=2$ GCN layers with $2,048$ hidden and $1,024$ output units. We use $0.5$ dropout regularization \cite{Srivastava:2014:DSW:2627435.2670313}\footnote{The code of our experiments is available at:
\url{https://github.com/druv022/Disease-Normalization-with-Graph-Embeddings}.}.

\begin{table}[tb]
    \centering
    \caption{Results on test set and validation set of the NER experiments. We use \cite{DBLP:journals/corr/LampleBSKD16} and \cite{DBLP:journals/corr/MaH16} models as a baseline, and compare with Habibi's et al. \cite{Maryam2017deep} known re-implementation of \cite{DBLP:journals/corr/LampleBSKD16}. For bioELMo we use the same architecture common to both the baseline. Results reported are the mean and standard deviation of 5 runs. *Our implementation. **Result taken from original paper.}
    \begin{tabular}{|p{0.34\textwidth}|c|c|c|c|}
        \hline
         Encoder & Pre & Rec & F1 & Val.~F1\\
        \hline
        Lample et al* & $0.824 \pm 0.022$ & $0.742 \pm 0.019$ & $0.781 \pm 0.003$ & $0.805 \pm 0.007$ \\
        \hline
        Ma and Hovy* & $0.823 \pm 0.011$ & $0.776 \pm 0.023$ & $0.799 \pm 0.012$ & $0.792 \pm 0.005$ \\
        \hline
        bioELMo & $0.878 \pm 0.003$ & $0.856 \pm 0.005$ & $0.867 \pm 0.002$ & $0.884 \pm 0.001$\\
        \hline\hline
        Lample et al.** \cite{Maryam2017deep} & $0.875$** & $0.836$** & $0.844$** & - \\
        \hline
    \end{tabular}
    \label{tab:NER_results}
\end{table} 
\begin{table}[tb]
    \centering
    \caption{Results on test set and validation set of our EL models for different type of MeSH encoding. Results reported are the mean and standard deviation of 5 runs. We consider unlexicalized \emph{node2vec} as our baseline. *Result taken from original paper. **We cannot compare fully due to different methodology. ``S.N" refers to Scope Note.}
    \begin{tabular}{|p{0.18\textwidth}|c|c|c|c|c|}
        \hline
    Disease emb. & MRR  & F1 & Pre & Pre@30& Val.~MRR\\
        \hline
        bioELMo (S.N.) & $0.748 \pm 0.002$ & $0.715 \pm 0.004$ & $0.715 \pm 0.002$ & $0.844 \pm 0.004$ &$0.791 \pm 0.001$\\
        \hline
        \nodeTvec \rom{1} & $0.749 \pm 0.002$ & $0.718 \pm 0.004$ & $0.720 \pm 0.004$  & $0.819 \pm 0.006$ & $0.800 \pm 0.003$ \\
        \hline
        \nodeTvec \rom{2} & $0.757 \pm 0.001$ & $0.721 \pm 0.004$ & $0.724 \pm 0.001$  & $0.842 \pm 0.004$ & $0.804 \pm 0.006$ \\
        \hline
        GCN & $0.744 \pm 0.006$ & $0.710 \pm 0.008$ & $0.710 \pm 0.007$ & $ 0.831 \pm 0.005$ & $0.803 \pm 0.007$\\
        \hline
        \hline
        DNorm* \cite{10.1093/bioinformatics/btt474} & - & 0.782** & - & - & -\\
        \hline
        NormCo* \cite{2019normco} & - & 0.840** & 0.878** & - & -\\
        \hline
    \end{tabular}
    \label{tab:EL_results}
\end{table}

\myparagraph{Metrics and Results}
%
We report precision (Pre), recall (Rec) and (micro averaged) F1 scores (F1) for NER. For EL it is customary to test whether the target canonical entity (a \mesh concept identifier) occurs among the top $k$ predictions \cite{2019normco}. Hence we report precision at confidence rank $k$ (Pre@k) and mean reciprocal rank (MRR) defined as $\text{MRR} = 1/|E| \cdot \sum \{1/\text{rank}_i \mid 1\leq i \leq |E|\}$. We use early stopping for model selection based on performance on the validation set, and report the average result of $5$ independent runs on the test and validation sets.

Table \ref{tab:NER_results} shows the results for NER. By replacing the word and character embeddings altogether with bioELMo embeddings, the results improve by a large margin. Increasing the number of parameters in the models did not affect model performance. Furthermore, we improve on \cite{Maryam2017deep}'s re-implementation of Lample et al.'s biLSTM-CRF model on the NCBI corpus (+$0.028$ F1 score points).

We tested our EL baseline---\nodeTvec type-\rom{1}, encoding only the \mesh structural information---against disease embeddings generated by averaging the bioELMo embedding of Scope Notes (see Table \ref{tab:EL_results}), and found it to perform better except on Pre@30. Node lexicalization, \nodeTvec type-\rom{2}, yields results on a par or better than the bioELMo baseline. Using GCN for training, we do not achieve any improvement. This suggests that both structural and lexical information in \mesh are important for normalization, and that taxonomy encoding is best achieved when computed independently from the EL task. This strategy seems on the other hand to underperform other normalization methods reported in the literature \cite{2019normco,10.1093/bioinformatics/btt474}. Our systems however do not incorporate many of their optimizations such as: normalization to disease synonym rather than \mesh identifier, abbreviation resolution, re-ranking and filtering, or coherence models.

Table \ref{tab:mtl_results} shows, finally, that using MTL results in an improved NER score ($+0.009$ F1 score points w.r.t.\ our best NER model in Table~\ref{tab:NER_results}). This indicates a certain degree of transfer from the EL to the NER task. It also yields a improvement w.r.t.\ similar NER and MTL for the NCBI corpus \cite{Maryam2017deep,10.1093/bioinformatics/bty869} ($+0.028$ points for NER and $+0.015$ for MTL).

\begin{table}[tb]
    \centering
    \caption{Results on test set of our MTL experiment where we report precision, recall and F1-score for NER task, and for EL we report MRR and precision. Results reported are the mean and standard deviation of 3 runs. *Result taken from oginal paper. 
    }
    \begin{tabular}{|p{0.145\textwidth}|@{}c@{}|@{}c@{}|@{}c@{}|@{}c@{}|@{}c|}
        \hline
              & \multicolumn{3}{c|}{NER} & \multicolumn{2}{c|}{EL} \\ \cline{2-6}
        Model & Pre  & Rec  & F1  & MRR & Pre@30\\
        \hline
        NER\,\&\,GCN & $0.880 \pm 0.003$ & $0.872 \pm 0.008$ & $0.876 \pm 0.003$ & \,$0.747 \pm 0.003$\, & \,$0.816 \pm 0.006$ \\
        \hline
        NER & $0.875 \pm 0.006$ & $0.869 \pm 0.001$ & $0.872 \pm 0.003$ & - & - \\
        \hline
        GCN & - & - & - & \,$0.745 \pm 0.001$\, & \,$0.816 \pm 0.001$ \\
        \hline\hline
        Wang* \cite{10.1093/bioinformatics/bty869} & \,$0.857 \pm 0.009$* & \,$0.864 \pm 0.004$* & \,$0.861 \pm 0.003$* & - & -\\
        \hline
    \end{tabular}\\
    \label{tab:mtl_results}
\end{table}

\myparagraph{Error analysis}
%
Regarding NER, most errors arise when dealing with multi-token entities: e.g., for \emph{sporadic T-PLL} only the head \emph{T-PLL} is detected. On the other hand, the bioELMo model is able to identify abbreviated disease names much better than the two baselines: e.g., \emph{T-PLL} occurring alone is always correctly detected. This might be because bioELMo starts at character-level and gradually captures token- and phrase-level information.

Our best EL model makes errors by confusing diseases with their \mesh ancestors and neighbors: e.g., it confuses \emph{D016399- Lymphoma, T-Cell} with \emph{D015458- Leukemia, T-Cell}, with which it shares an ancestor: \emph{D008232- Lymphoproliferative Disorders}. Often, it resolves correctly the first instance of a concept but returns instead a neighbour afterwards: e.g., in ``\emph{Occasional missense mutations in ATM were also found in tumour DNA from patients with B-cell non-Hodgkins lymphomas (B-NHL) and a B-NHL cell line}" \emph{B-cell non-Hodgkins lymphomas} is correctly resolved to \emph{D016393} but \emph{B-NHL} (its abbreviation) is mapped to \emph{D008228} (its child) and the second occurrence to \emph{D018239} (another form of cancer). This might be due to the following facts: Both NCBI and \mesh are not particularly large resources, hence the mention and disease encodings learnt by the model are not sufficiently discriminative. In addition, GCN and \nodeTvec training tends to ignore the direction of \mesh taxonomical edges. Finally, DNorm and NormCo normalize diseases to concepts indirectly via their synonyms, allowing them to exploit better lexical features\footnote{They also include a wide range of optimizations such as re-ranking, coherence models or abbreviation resolution.}.

\section{Conclusions}

In this paper, we address the problem of normalizing disease names to \mesh concepts (canonical identifiers) by adapting state-of-the-art neural graph embeddings (GCN and \nodeTvec) that exploit both \mesh's hierarchical structure and the description of diseases. Our graph-based disease node encoding is the first of its kind, to best of our knowledge. We also apply multi-tasking to transfer information between disease detection (NER) and resolution (EL) components of the task, leveraging their common signals to improve on the single models. We observe that bioELMo embeddings lead to substantial improvement in NER performance. We demonstrate that node lexicalization does improve over either pure structural or lexical embeddings, and that MTL gives rise to state-of-the-art performance for NER in the corpus used (NCBI corpus).

On the other hand, we do not currently outperform other disease normalization approaches \cite{2019normco,10.1093/bioinformatics/btt474}, and often confuse neighbour \mesh concepts with their true targets. In the future we intend to experiment with the optimizations reported (especially: resolution to synonym rather than concept identifier and re-ranking) by \cite{2019normco,10.1093/bioinformatics/btt474}. \mesh is also a comparatively small graph with short disease descriptions. As further work we plan to enrich it by linking it to larger scale resources (e.g., Wikipedia).

\bibliographystyle{plain}
\bibliography{references}

\end{document}